\documentclass[letterpaper]{article} % DO NOT CHANGE THIS
\usepackage{aaai2026}  % DO NOT CHANGE THIS
\usepackage{times}  % DO NOT CHANGE THIS
\usepackage{helvet}  % DO NOT CHANGE THIS
\usepackage{courier}  % DO NOT CHANGE THIS
\usepackage[hyphens]{url}  % DO NOT CHANGE THIS
\usepackage{graphicx} % DO NOT CHANGE THIS
\usepackage{natbib}  % DO NOT CHANGE THIS AND DO NOT ADD ANY OPTIONS TO IT
\usepackage{caption} % DO NOT CHANGE THIS AND DO NOT ADD ANY OPTIONS TO IT
\usepackage{algorithm}
\usepackage{algorithmic}
\usepackage{booktabs}
\usepackage{amsmath}

\usepackage{newfloat}
\usepackage{listings}
\DeclareCaptionStyle{ruled}{labelfont=normalfont,labelsep=colon,strut=off} % DO NOT CHANGE THIS
\floatstyle{ruled}
\newfloat{listing}{tb}{lst}{}
\floatname{listing}{Listing}
\title{An Interpretability-Guided Framework for Responsible Synthetic Data Generation in Emotional Text}
\author{
    Paula Joy B. Martinez,
    Jose Marie Antonio Miñoza,
    Sebastian C. Ibañez
}
\affiliations{
    Center for AI Research (CAIR)\\
    Department of Education, Philippines
}

\usepackage{bibentry}
\begin{document}

\maketitle

\begin{abstract}

Emotion recognition from social media is critical for understanding public sentiment, but accessing training data has become prohibitively expensive due to escalating API costs and platform restrictions.
We introduce an interpretability-guided framework where Shapley Additive Explanations (SHAP) provide principled guidance for LLM-based synthetic data generation.
With sufficient seed data, SHAP-guided approach matches real data performance, significantly outperforms naïve generation, and substantially improves classification for underrepresented emotion classes.
However, our linguistic analysis reveals that synthetic text exhibits reduced vocabulary richness and fewer personal or temporally complex expressions than authentic posts.
This work provides both a practical framework for responsible synthetic data generation and a critical perspective on its limitations, underscoring that the future of trustworthy AI depends on navigating the trade-offs between synthetic utility and real-world authenticity.
\end{abstract}


\section{Introduction}

\begin{figure*}[t]
\centering
\includegraphics[width=\textwidth]{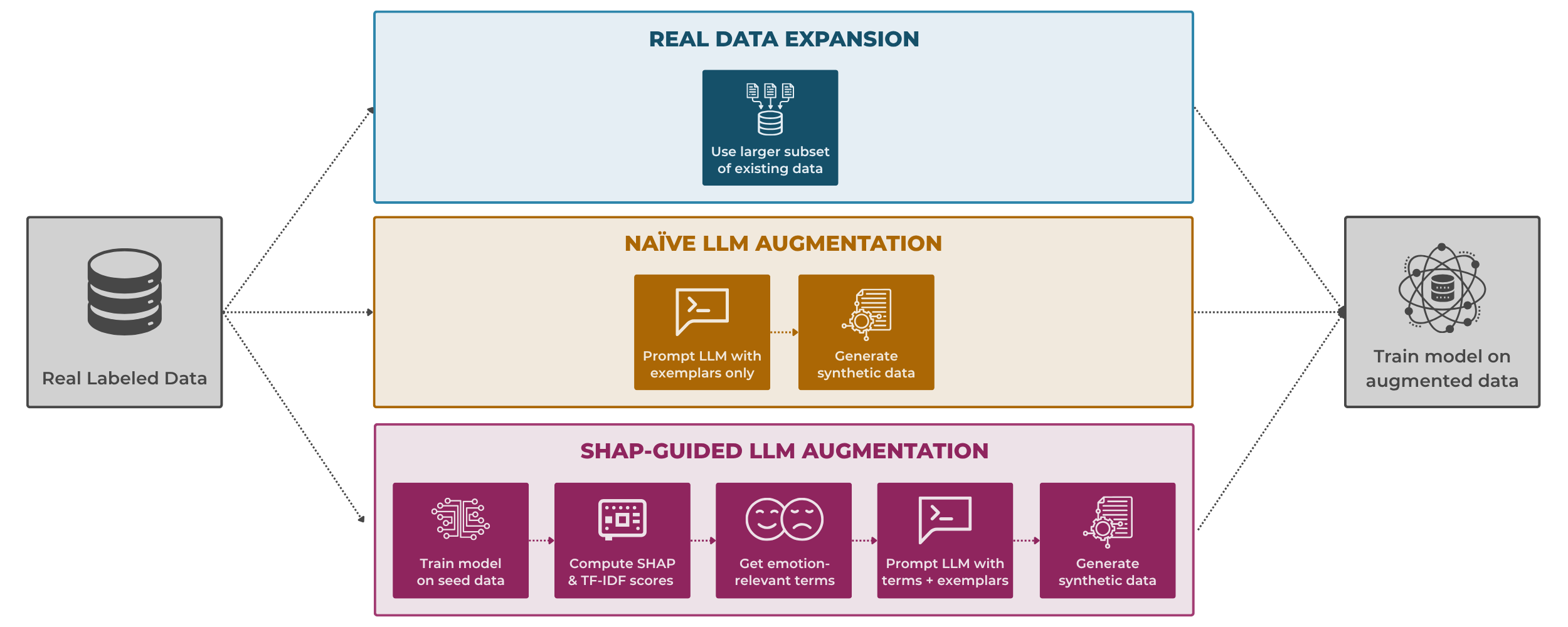}
\caption{The experimental setup comparing three augmentation strategies: real data expansion, SHAP-guided generation (exemplars + SHAP keywords), and naïve generation (exemplars only). SHAP analysis uses the baseline model to extract emotion-specific keywords. 
All strategies are evaluated with identical model parameters, data splits, 
and incremental testing.}
\label{fig:overall_diagram}
\end{figure*}

Emotion recognition from social media has become critical for understanding public sentiment across applications including political forecasting, policy analysis, and social movement detection~\cite{Forciniti2023}. 
However, acquiring training data has become prohibitively expensive. 
Major platforms like X (formerly Twitter), TikTok, and Reddit imposed restrictive API changes throughout 2023~\cite{UniversityofBath2023, 
Davidson2023, Mehta2024}, while projections indicate LLM training datasets will exhaust available public text between 2026 and 2032~\cite{Villalobos2022}. 
These constraints intensify the need for augmentation techniques that generate high-quality emotional text without continuous platform access.

Traditional rule-based methods are efficient and interpretable, but limited to short transformations~\cite{Shorten2021} that fail to capture social media's linguistic nuances. While neural~\cite{Sennrich2016, Kumar2019} and LLM-based methods~\cite{AnabyTavor2020, Zhao2023, Whitehouse2023, Uveges2025} improved sample quality, naïve generation lacks transparency and quality control, which motivated research on principled prompt design~\cite{Yoo2021, Wang2022, Meng2022, Li2023, Jain2025}.

Despite these advances, explainable AI techniques remain underexplored for guiding responsible synthetic data generation.
In particular, Shapley Additive Explanations (SHAP)~\cite{Lundberg2017} is widely used in classification tasks~\cite{Kong2022, Hailemariam2023, Kim2024} but its role in LLM data augmentation for emotion recognition remains insufficiently studied.
Moreover, most prior work applies interpretability only indirectly through post-hoc filtering~\cite{Wu2022} and feature modification~\cite{Mersha2025}, and no prior work has directly integrated SHAP-derived feature importance into prompt construction for text generation.

We address this gap by using token-level SHAP scores to guide LLMs toward generating emotionally aligned synthetic text, particularly in low-resource settings.
Our contributions include: (1) an interpretability-guided augmentation framework (Figure~\ref{fig:overall_diagram}) that conditions LLM generation on feature importance; (2) an evaluation of three augmentation strategies––real data expansion, SHAP-guided generation, and naïve generation––across varying data regimes; (3) and a linguistic analysis discussing a fundamental trade-off, where synthetic data exhibits reduced lexical diversity but amplified emotion-discriminative features that benefit minority class performance.

\section{Methods and Experimental Design}

We evaluated six traditional ML algorithms using the TweetEval dataset (Appendix~\ref{app:dataset}): K-Nearest Neighbors, Random Forest, XGBoost, AdaBoost, HistGradientBoosting, and LightGBM.
Using real data expansion as the baseline augmentation strategy, we tested each classifier across all data increments and over several folds to identify the algorithm that would provide the most reliable performance benchmarks.
XGBoost consistently demonstrated the highest mean F1-score (Table~\ref{tab:model_selection}) while maintaining stable performance patterns throughout the augmentation process. 
This stability was crucial to ensure that the observed performance differences between augmentation strategies could be attributed to their effectiveness rather than to model-specific inconsistencies.
We therefore used XGBoost exclusively for all subsequent experiments.

\subsection{Data Augmentation Strategies}

\subsubsection{Real Data Expansion}

Real data expansion represents the traditional approach to improving model performance through the acquisition of additional manually labeled data. 
The candidate real samples are the remaining observations after the removal of the holdout set and the seed data. 
This strategy serves as the baseline for evaluating whether synthetic data can provide comparable benefits to authentic data while avoiding the resource constraints associated with manual annotation and extraction costs.

\subsubsection{Interpretability-guided Synthetic Data Generation}

This method combines Shapley analysis with TF-IDF differential scoring to identify emotion-relevant keywords.
We trained the classifier on the seed data and applied SHAP to quantify feature importance. 
We then calculated the mean TF-IDF for each feature within a target emotion and subtracted the mean TF-IDF values of all other emotions (e.g., for anger, we compute average TF-IDF for anger tweets minus the average for joy, optimism, and sadness combined).
This differential scoring reveals which terms are discriminatively associated with specific emotions, where positive differences indicate words more frequent in the target emotion and negative differences highlight words more characteristic of other emotions.

We filtered these terms by their SHAP importance scores to exclude frequent but non-predictive words, ensuring the selected keywords are not just common but are also what the model considers important in predicting emotions. 
The remaining keywords are then ranked and split into positive/negative sets (examples in Appendix~\ref{app:keywords}).
Each generation request combines real Tweet exemplars with SHAP-derived keywords, and the LLM is instructed to naturally incorporate positive keywords and avoid negative ones (prompt template in Appendix~\ref{app:prompt}).
Ablation studies (Appendix~\ref{app:ablation}) demonstrate that both components are essential for optimal performance.

\subsubsection{Naïve Synthetic Generation}

The naïve approach employs the same synthetic data generation pipeline as the SHAP-guided approach, but without using insights from SHAP. 
The LLM still receives the same set of real Tweet exemplars for each emotion, but unlike the SHAP-guided approach, the prompt contains no specific keyword instructions about which features to emphasize or avoid. 
Instead, the LLM relies entirely on its pre-trained understanding of emotional language and the patterns it can infer from the provided exemplars to generate emotionally appropriate samples. 
By comparing model performance when trained with naïve versus SHAP-guided synthetic data, we determine whether the additional complexity of keyword extraction and prompt engineering indeed translates into measurable improvements in emotion recognition.

\subsection{Computational Linguistic Analysis}

Performance improvements alone do not shed light on the linguistic characteristics of the generated data, and they raise a fundamental question: does the model glean generalizable emotional patterns or does it merely memorize the linguistic characteristics of real data?
This distinction is crucial for responsible synthetic data generation in emotion recognition, where apparent gains may reflect overfitting to artifacts rather than genuine linguistic understanding. 
To assess both fidelity and utility of LLM-generated data, we evaluate augmentation effectiveness through classification performance (F1-scores) and linguistic diversity analysis.
For lexical diversity, we measure Type-Token Ratio (TTR) to quantify vocabulary richness, and Jaccard Index to assess vocabulary overlap between synthetic and real data.
For syntactic diversity, we analyze Part-of-Speech (POS) bigram and trigram patterns using Jensen-Shannon Divergence (JSD) to quantify structural differences between LLM-generated and authentic social media text.
Together, these quality metrics contribute to evaluation standards for LLM-generated training data (detailed formulas in Appendix~\ref{app:linguistic_analysis}). 

\subsection{Experimental Setup}

We split the TweetEval dataset into a holdout set (approx. 1,000 samples) and training pool (approx. 4,000 samples). 
To evaluate conditional effectiveness across data regimes, we conduct experiments with three seed sizes: 100, 500, and 1,000 samples. 
For each seed size, we progressively add data in fixed increments (+100, +200, etc.) and train the model on each augmented set, then evaluate on the holdout set via cross-validation.

Using the baseline model and seed data, we perform SHAP analysis to extract keywords for guided generation. 
Two synthetic datasets are generated in parallel: SHAP-guided (exemplars + keywords) and naïve (exemplars only), then evaluated using identical splits and incremental testing as the real data baseline.

\section{Results and Analysis}

\begin{figure}[!ht]
\includegraphics[width=\columnwidth]{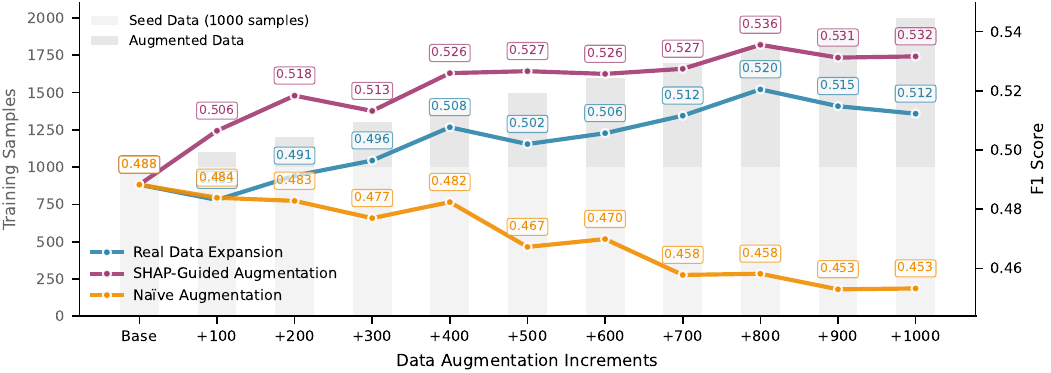}
\caption{SHAP-guided synthetic data achieves equivalent performance to real data expansion, while naïve synthetic data degrades consistently, highlighting the importance of principled generation approaches.}
\label{fig:1000_XGBoost_all}
\end{figure}

Figure~\ref{fig:1000_XGBoost_all} demonstrates that SHAP-guided augmentation achieves performance equivalent to real data expansion (F1: 0.48-0.53) across all increments from 1,000 to 2,000 samples.
Conversely, naïve augmentation shows consistent degradation (0.48→0.45), which demonstrates that synthetic data generation requires quality control: without interpretability guidance, generated data introduces noise that degrades rather than improves model performance.

\begin{figure}[!ht]
\centering
\includegraphics[width=\columnwidth]{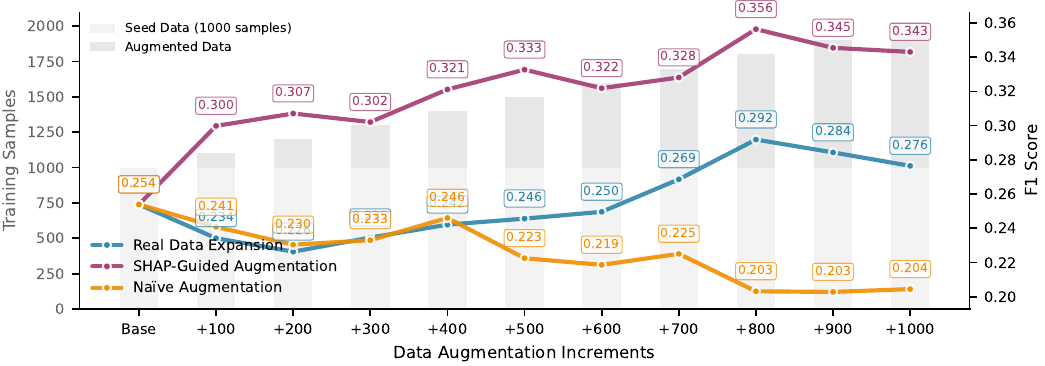}
\caption{SHAP-guided augmentation excels for minority class optimism (8.8\%), demonstrating how interpretability-guided generation can address class imbalance.}
\label{fig:optimism}
\end{figure}

For optimism (8.8\% of data), SHAP-guided augmentation consistently outperforms both alternatives (Figure~\ref{fig:optimism}), demonstrating particular effectiveness for minority 
classes where focused exposure to discriminative patterns is critical. 
This addresses a key fairness concern, wherein synthetic data can enhance representation for underrepresented classes when guided by interpretability insights.

\begin{figure}[!ht]
\centering
\includegraphics[width=\columnwidth]{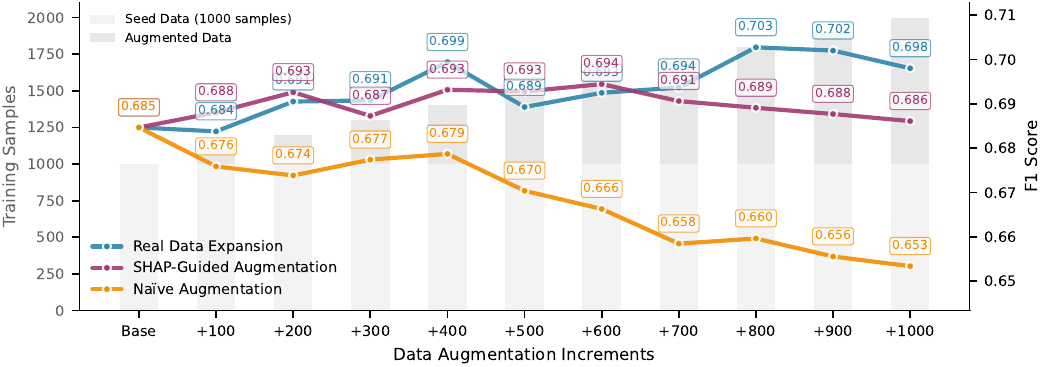}
\caption{All strategies show stable performance for majority class anger (42\%), demonstrating that adequate baseline data reduces augmentation sensitivity.}
\label{fig:anger}
\end{figure}

For anger (42\% of data), all strategies show stable performance with marginal differences (Figure~\ref{fig:anger}), indicating that adequate training data reduces sensitivity to augmentation quality. 
This frequency-performance relationship reveals a clear pattern where SHAP-guided augmentation provides maximal value for minority classes with limited authentic examples, while majority classes show robustness across strategies.
Similar patterns emerged for sadness and joy (Appendix~\ref{app:emotions}), with SHAP-guided approaches maintaining performance comparable to real data while naïve generation degraded consistently.

\subsection{Model Performance on Different Data Regimes}

\begin{figure}[h]
\includegraphics[width=\columnwidth]{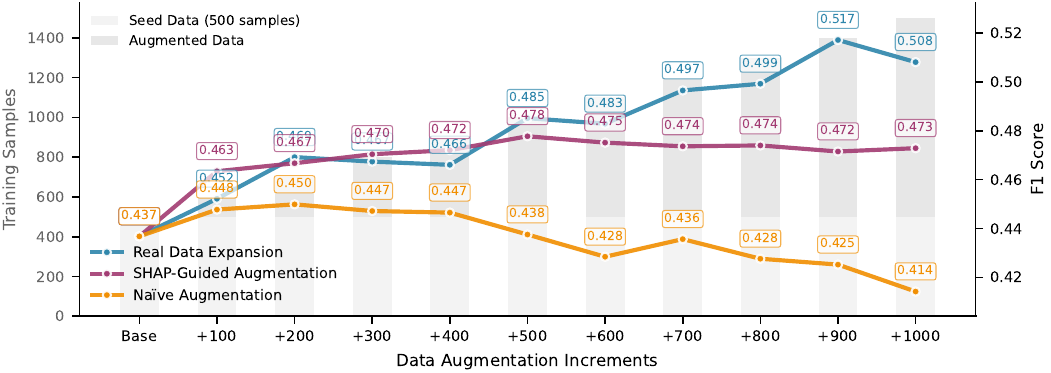}
\caption{With 500-sample baseline, SHAP-guided generation achieves parity with real data initially but plateaus at higher increments, demonstrating that interpretability-guided generation requires adequate seed data for sustained effectiveness.}
\label{fig:500_XGBoost_all}
\end{figure}

Figure \ref{fig:500_XGBoost_all} shows a shift in the effectiveness of SHAP-guided augmentation under reduced seed data. 
At the 500-sample baseline, SHAP-guided augmentation closely aligns with real data expansion and consistently outperforms naïve augmentation across all increments.
However, its gains taper off with larger increments and eventually falls behind real data, unlike in the 1,000-sample baseline (Figure \ref{fig:1000_XGBoost_all}), where SHAP-guided data consistently leads.

\begin{figure}[h]
\centering
\includegraphics[width=\columnwidth]{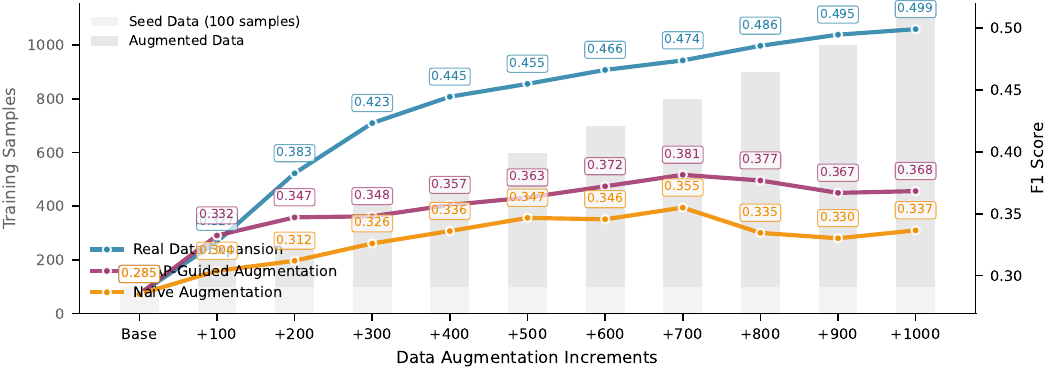}
\caption{At 100-sample baseline, SHAP-guided generation marginally outperforms naïve but remains below real data throughout, demonstrating that interpretability guidance requires minimally sufficient seed data to provide any advantage.}
\label{fig:100_XGBoost_all}
\end{figure}

This downward trend becomes even more pronounced at the 100-sample baseline (Figure \ref{fig:100_XGBoost_all}), where SHAP-guided augmentation offers only marginal improvement over naïve generation and remains consistently below real data.
These findings underscore that while interpretability-guided generation is highly effective, its performance depends on a minimally sufficient dataset to extract meaningful emotional patterns. 
When this threshold is not met, its advantage diminishes. 
Thus, even with interpretability, LLM-based augmentation complements rather than replaces real data in extremely low-resource scenarios.

\subsection{Linguistic Analysis of Synthetically Generated Emotional Text}

Both the SHAP-guided and naïve approaches exhibit lower lexical diversity than real data (TTR: SHAP 0.133, Naïve 0.143 vs. Real 0.241, Table~\ref{tab:linguistic_metrics}), with SHAP-guided showing lowest diversity due to focused repetition of emotion-discriminative features. 
This concentration amplifies salient patterns beneficial for minority classes but risks overfitting to specific constructions, and is particularly problematic for social media's evolving linguistics.

\begin{table}[h]
\centering
\caption{Lexical diversity (TTR) and overlap (Jaccard) metrics. TTR measures vocabulary richness within each dataset (higher = more diverse).
Jaccard measures lexical overlap between real and synthetic data (higher = more similar vocabularies).}
\label{tab:linguistic_metrics}
\begin{tabular}{@{}lccc@{}}
\toprule
\textbf{Metric} & \textbf{Real Data} & \textbf{SHAP-Guided} & \textbf{Naïve} \\
\midrule
Type-Token Ratio & 0.241 & 0.133 & 0.143 \\
\midrule
\multicolumn{4}{l}{\textit{Jaccard Similarity (vs. Real Data):}} \\
Anger    & --- & 0.209 & 0.184 \\
Optimism & --- & 0.247 & 0.183 \\
Sadness  & --- & 0.232 & 0.210 \\
Joy      & --- & 0.188 & 0.171 \\
\bottomrule
\end{tabular}
\end{table}

\begin{figure}[t]
\includegraphics[width=\columnwidth]{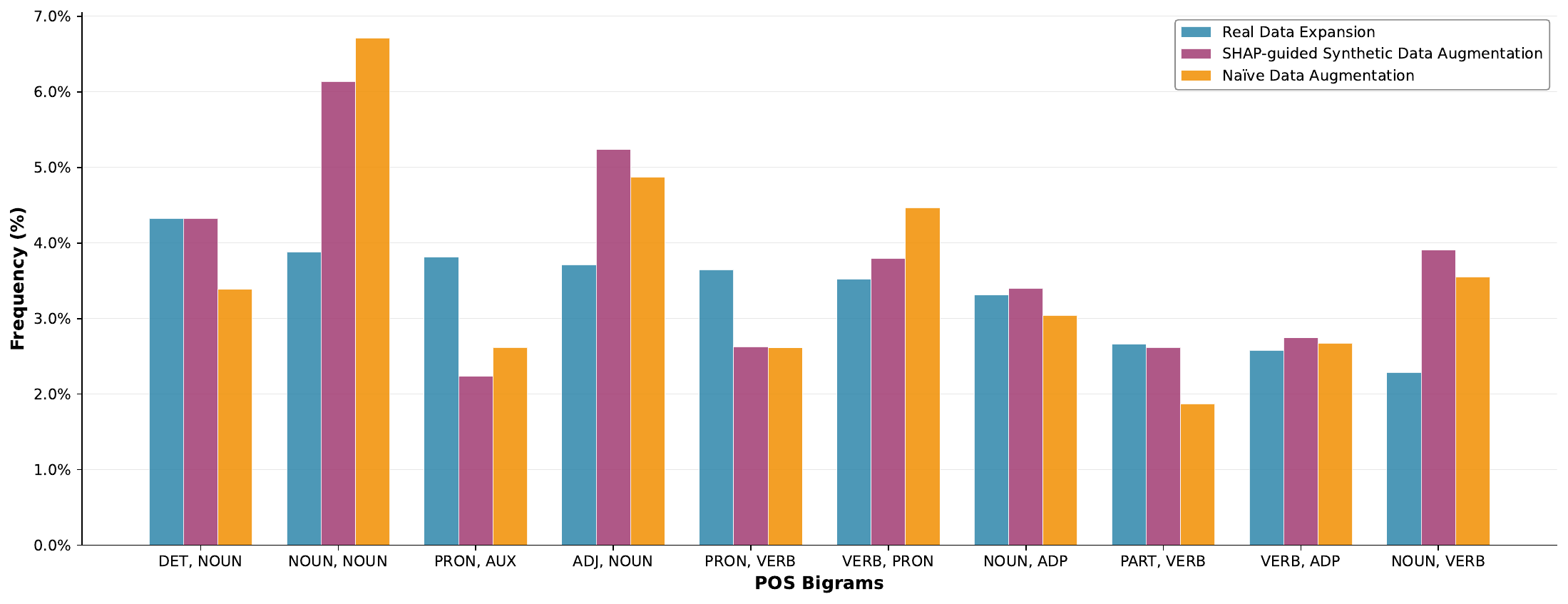}
\caption{POS bigram frequencies showing SHAP-guided emphasis on content-rich patterns while underrepresenting personal voice constructions.}
\label{fig:bigram}
\end{figure}

SHAP-guided generation also achieves higher lexical overlap with real data across all emotions (Jaccard: 0.188-0.247 vs. naïve: 0.171-0.210), with optimism showing strongest alignment despite being the minority class. Syntactic analysis (Figure~\ref{fig:bigram}) further reveals that SHAP-guided data overproduces content-rich patterns (ADJ-NOUN, NOUN-VERB) that convey clearer emotional cues, but both synthetic approaches underproduce personal voice patterns (PRON-AUX, PRON-VERB), reducing first-person perspectives that are characteristic of authentic social media text (extended analysis and examples in Appendix~\ref{app:linguistic}).
These findings demonstrate that SHAP-guided generation succeeds not by broadly mimicking natural language, but by amplifying emotion-discriminative features—achieving performance gains through focused repetition that produces emotionally authentic but lexically homogeneous text. This trade-off between effectiveness and diversity implies that, while synthetic data efficiently addresses immediate classification needs, real data remains essential for capturing the linguistic variety required for robust generalization across evolving social media contexts.

\section{Conclusions and Future Work}

Our work demonstrates that synthetic data augmentation, when guided by interpretability, offers a viable solution for emotion recognition in data-constrained scenarios, but only under specific conditions that must inform responsible deployment.
SHAP-guided generation achieves performance parity with real data expansion for underrepresented classes when adequate seed data enables meaningful feature extraction and interpretability insights are systematically incorporated.
However, this effectiveness comes with critical trade-offs: while SHAP-guided augmentation amplifies discriminative emotional patterns beneficial for classification, it produces lexically homogeneous text that substantially reduces linguistic diversity compared to real data.

These findings establish clear guardrails for responsible synthetic data use in emotion recognition. First, practitioners must ensure minimum seed data thresholds before deploying interpretability-guided generation. Second, the reduced linguistic diversity of synthetic data necessitates continued real data acquisition to prevent overfitting to repetitive patterns and maintain generalization across evolving linguistic contexts, demographic variations, and platform-specific norms.
Future work should extend these findings to transformer-based architectures, examine cross-platform transferability where linguistic norms differ substantially, and develop automated quality metrics that detect when synthetic data degrades rather than enhances model fairness.

\bibliography{aaai2026}

\section*{Appendix}
\appendix
\setcounter{secnumdepth}{2}

This appendix provides supplementary materials supporting the analyses presented in the main text, including detailed model comparisons, extended emotion-specific results, comprehensive linguistic diversity analysis, and ablation studies.

%% ====== APPENDIX: MODEL SELECTION ======
\renewcommand{\thefigure}{A\arabic{figure}}
\renewcommand{\thetable}{A\arabic{table}}
\setcounter{figure}{0}
\setcounter{table}{0}
\section{Initial Model Training}
\label{app:model}

We compared six traditional machine learning algorithms using real data expansion as the baseline. Table~\ref{tab:model_selection} demonstrates XGBoost's consistent superiority and stable performance across all augmentation increments and cross-validation folds.

\begin{table}[h]
\centering
\caption{Classifier performance across 10-fold cross-validation with 10 data increments (+100 to +1,000), sorted by mean F1-score}
\label{tab:model_selection}
\begin{tabular}{@{}lccc@{}}
\toprule
\textbf{Classifier} & \textbf{Mean F1} & \textbf{Std Dev} & \textbf{CV (\%)} \\
\midrule
Random Forest            & 0.4321 & 0.1064 & 24.6 \\
\textbf{XGBoost}         & \textbf{0.4307} & \textbf{0.0729} & \textbf{16.9} \\
KNN                      & 0.2883 & 0.0787 & 27.3 \\
HistGradientBoosting     & 0.2454 & 0.0541 & 22.0 \\
LightGBM                 & 0.2444 & 0.0531 & 21.7 \\
AdaBoost                 & 0.2280 & 0.0306 & 13.4 \\
\bottomrule
\end{tabular}
\end{table}

%% ====== APPENDIX: TweetEval Dataset ======
\setcounter{figure}{0}
\setcounter{table}{0}
\renewcommand{\thefigure}{B\arabic{figure}}
\renewcommand{\thetable}{B\arabic{table}}
\section{TweetEval Dataset}
\label{app:dataset}

This paper uses the TweetEval dataset~\cite{Barbieri2020}, a repository consisting of seven heterogeneous Twitter-specific classification tasks, one of which is emotion recognition (Table~\ref{tab:data_sample}). It comprises 5,052 tweets with no missing values and has an imbalanced class distribution across four emotional categories. The preprocessing pipeline was designed to preserve emotionally relevant linguistic patterns while removing noise commonly found in social media data. 
Raw tweets were rid of user mentions, hashtag symbols, and non-alphanumeric characters. 
Emotion-bearing linguistic elements were also retained, such as negation markers ("not", "never", "can't"), intensity modifiers ("very", "extremely", "quite"), and explicit emotion terms.
The term frequency-inverse document frequency (TF-IDF) vectorization was then applied to highlight words that are distinct to a particular document.

\begin{table}[t]
\centering
\caption{Sample tweets from TweetEval emotion recognition dataset illustrating the four emotion classes with typical social media characteristics.}
\label{tab:data_sample}
\small
\begin{tabular}{@{}p{0.55\columnwidth}cc@{}}
\toprule
\textbf{Text} & \textbf{Label} & \textbf{Emotion} \\ 
\midrule
My roommate: it's okay that we can't spell because we have autocorrect. \#terrible \#firstworldprobs & 0 & anger \\
\addlinespace
No but that's so cute. Atsu was probably shy about photos before but cherry helped her out uwu & 1 & joy \\
\addlinespace
Worry is a down payment on a problem you may never have & 2 & optimism \\
\addlinespace
it's pretty depressing when u hit pan on ur fave eyeshadow & 3 & sad \\
\bottomrule
\end{tabular}
\end{table}

%% ====== APPENDIX: SHAP KEYWORDS ======
\section{SHAP-Derived Keywords}
\setcounter{figure}{0}
\setcounter{table}{0}
\renewcommand{\thefigure}{C\arabic{figure}}
\renewcommand{\thetable}{C\arabic{table}}
\label{app:keywords}

Table~\ref{tab:topkeywords} shows the top five keywords per emotion class extracted 
using differential TF-IDF scoring and filtered by SHAP importance. These keywords 
were incorporated into the generation prompts to guide the LLM toward 
emotion-specific linguistic patterns.

\begin{table}[h]
\centering
\caption{Top 5 keywords per emotion extracted using differential TF-IDF and SHAP 
filtering to identify terms that are both frequent and predictive.}
\label{tab:topkeywords}
\begin{tabular}{p{0.21\columnwidth}p{0.21\columnwidth}p{0.21\columnwidth}p{0.21\columnwidth}}
\hline
\textbf{Anger} & \textbf{Joy} & \textbf{Optimism} & \textbf{Sadness} \\
\hline
angry, fuming, outrage, anger, bully & love, blues, happy, new, hilarious, birthday & 
worry not, fears, start, optimism & sadness, depression, lost, so depressing \\
\hline
\end{tabular}
\end{table}

%% ====== APPENDIX: PROMPT TEMPLATE ======
\section{LLM Generation Prompt Template}
\setcounter{figure}{0}
\setcounter{table}{0}
\renewcommand{\thefigure}{D\arabic{figure}}
\renewcommand{\thetable}{D\arabic{table}}
\label{app:prompt}

Table~\ref{tab:prompt} presents the complete prompt template used for SHAP-guided synthetic data generation. 
The prompt explicitly requests realistic social media language patterns, including hashtags, user mentions, and casual expressions that mirror authentic Twitter posts.

All synthetic emotional text was generated using Claude 3.5 Sonnet (model: \texttt{claude-3-5-sonnet-20241022}) via the Anthropic API. 
Generation parameters were configured as follows: temperature = 0.8 to encourage lexical diversity while maintaining coherent emotional expression, max tokens = 1,500 per API call, and batch size = 20 tweets per generation request. 
A 2-second delay was implemented between API calls to respect rate limits.
The system prompt instructed the model to act as "an expert at generating realistic social media content" for the target emotion, with explicit requirements to produce diverse, authentic-sounding tweets incorporating hashtags, mentions (@user), casual expressions, and informal language patterns 
characteristic of social media discourse.

\begin{table*}[t]
\centering
\caption{LLM prompt template for SHAP-guided synthetic tweet generation incorporating 
real tweet exemplars and emotion-specific keywords with explicit instructions for 
authentic social media language patterns.}
\label{tab:prompt}
\begin{tabular}{p{0.95\textwidth}}
\hline
\begin{minipage}{0.93\textwidth}
\footnotesize
\begin{verbatim}
I need to generate {current_batch_size} realistic {emotion} tweets 
for a machine learning emotion detection dataset.

{exemplar_text ? "EXAMPLES OF REAL " + emotion.upper() + " TWEETS:\n" 
+ exemplar_text : ""}

{include_text ? "\nKEYWORDS TO INCLUDE (use these concepts naturally 
and randomly): " + include_text : ""}

{exclude_text ? "\nKEYWORDS TO AVOID (don't emphasize these): " 
+ exclude_text : ""}

REQUIREMENTS:
1. Generate exactly {current_batch_size} {emotion} tweets
2. Make them realistic social media posts with natural language
3. Use hashtags, mentions (@user), and casual expressions when 
   appropriate
4. Each tweet should clearly express {emotion}
5. Number each tweet (1., 2., 3., etc.)
6. Do not create files. Just give the tweet.
7. Do not repeat the themes of the tweet. Be imaginative and put 
   yourself in the circumstances of humans.
8. You are not allowed to give the same tweets that you have 
   already provided.
9. Sample keywords randomly and make some tweets informal, with 
   typos, slang

Generate the {emotion} tweets now:
\end{verbatim}
\end{minipage} \\
\hline
\end{tabular}
\end{table*}

%% ====== APPENDIX: COMPUTATIONAL LINGUISTIC ANALYSIS ======
\section{Computational Linguistic Analysis}\
\setcounter{figure}{0}
\setcounter{table}{0}
\renewcommand{\thefigure}{E\arabic{figure}}
\renewcommand{\thetable}{E\arabic{table}}
\label{app:linguistic_analysis}

\subsubsection{Lexical Diversity}

Lexical diversity is important for language models to generalize effectively across contexts. 
Diverse vocabulary enables a model to move beyond simple pattern recognition and develop deeper linguistic understanding. 
However, synthetic text often repeats common words and phrases rather than introducing novel expressions~\cite{Uveges2025}, and limited vocabulary in synthetic datasets can cause models to over-rely on familiar patterns, therefore reducing their ability to handle new textual inputs.

\paragraph{Type-Token Ratio (TTR)}
A common way to measure lexical diversity is through the Type-Token Ratio (TTR)~\cite{Kettunen2014, Xu2023}, which is formalized as follows:

\begin{equation}
    TTR = \frac{\text{Number of Unique Words (Types)}}{\text{Total Number of Words (Tokens)}}
\end{equation}

This metric ranges from 0 to 1, where 1 represents text with no repeated words. 
A higher TTR means the text uses more varied vocabulary, while a lower TTR shows that words are being repeated more often in the dataset. 
We calculated TTR across the augmented datasets produced by all three data augmentation strategies. 

\paragraph{Jaccard Index}
The Jaccard similarity coefficient~\cite{Hancock2004} is a popular measure of set similarity between two datasets $A$ and $B$. 
In this paper, we define the Jaccard Index as:
\begin{equation}
    J(A, B) = \frac{|A \cap B|}{|A \cup B|}
\end{equation}

where $A$ denotes the set of distinct words found in the real dataset and $B$ denotes the set of distinct words present in the artificial dataset generated via the SHAP-guided or naïve approach. 
Scores close to 1 reflect strong vocabulary alignment between two datasets and scores near 0 indicates substantial lexical divergence.

\subsubsection{Syntactic Analysis}
To assess grammatical differences between authentic and synthetic datasets, we examined Part-of-Speech (POS) patterns through bigram frequency analysis.
The Jensen-Shannon Divergence (JSD)~\cite{Lin1991} serves as our metric for quantifying the differences between POS n-gram distributions across datasets.
JSD provides a symmetric and normalized measure of divergence between probability distributions $P$ and $Q$, and is computed as:
\begin{equation}
    JSD(P \parallel Q) = \frac{1}{2} D_{KL}(P \parallel M) + \frac{1}{2} D_{KL}(Q \parallel M)
\end{equation}
where $M = \frac{1}{2}(P + Q)$ represents the average distribution and $D_{KL}$ indicates the Kullback-Leibler divergence. 
Higher JSD values signal greater structural differences in POS patterns, while lower values indicate more similar syntactic structures.

%% ====== APPENDIX: EMOTION-SPECIFIC ANALYSIS ======
\section{Emotion-Specific Performance Analysis}
\setcounter{figure}{0}
\setcounter{table}{0}
\renewcommand{\thefigure}{F\arabic{figure}}
\renewcommand{\thetable}{F\arabic{table}}
\label{app:emotions}

This section provides performance analysis for sadness and joy emotion 
classes to complement the optimism and anger results presented in the main text.

\subsection{Sadness Classification}

\begin{figure*}[h]
\centering
\includegraphics[width=\textwidth]{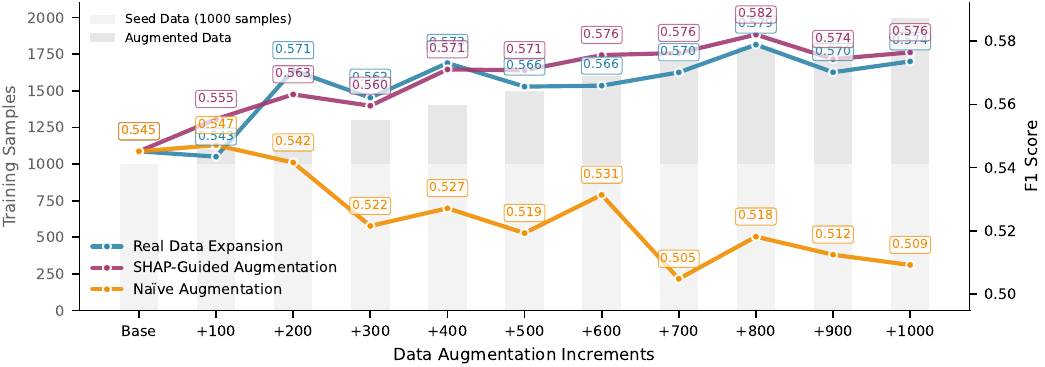}
\caption{XGBoost F1-score performance for sadness class across three augmentation strategies. 
SHAP-guided synthetic data maintains high performance comparable to real data expansion, while naïve synthetic data shows performance degradation. 
This highlights the effectiveness of interpretability-guided generation for capturing sadness-specific linguistic patterns.}
\label{fig:sad}
\end{figure*}

For the sadness class (Figure~\ref{fig:sad}), SHAP-guided augmentation shows the most pronounced advantage over other strategies. 
This approach maintains consistently high F1-scores between 0.537 and 0.576 throughout all increments, closely matching the performance of the real data expansion strategy (0.538-0.589). 
This suggests that SHAP-guided keywords effectively capture the linguistic patterns most discriminative for sadness classification. 
In contrast, naïve augmentation shows degradation from 0.526 to 0.515, suggesting that unguided synthetic data deviates significantly from authentic emotional patterns.

\subsection{Joy Classification}

\begin{figure*}[h]
\centering
\includegraphics[width=\textwidth]{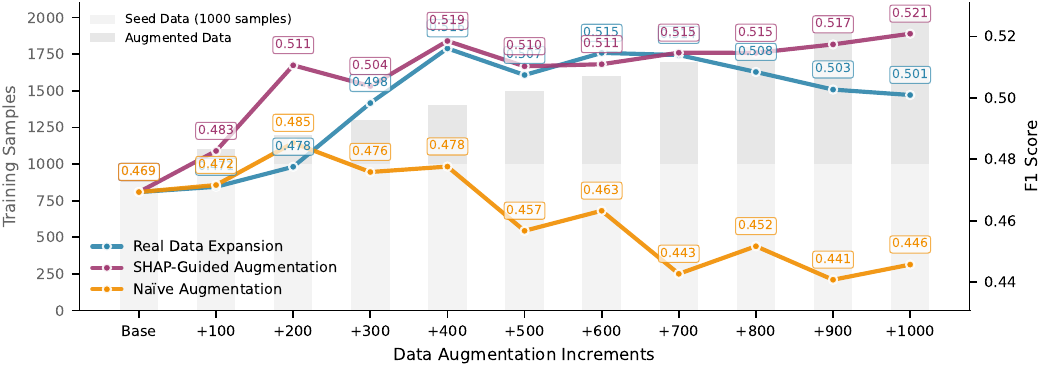}
\caption{F1-score performance for joy emotion classification showing SHAP-guided synthetic data achieving near-equivalent performance to real data expansion with a smaller performance gap compared to other emotion categories, while naïve 
synthetic data continues to underperform both strategies.}
\label{fig:joy}
\end{figure*}

The same pattern is observed in the joy class (Figure~\ref{fig:joy}), albeit the gap between real data expansion and SHAP-informed data generation is smaller. 
SHAP-guided synthetic data achieves near-equivalent performance to real data expansion, while naïve synthetic data continues to underperform both strategies.

%% ====== APPENDIX: DETAILED LINGUISTIC ANALYSIS ======
\section{Detailed Linguistic Analysis}
\setcounter{figure}{0}
\setcounter{table}{0}
\renewcommand{\thefigure}{G\arabic{figure}}
\renewcommand{\thetable}{G\arabic{table}}
\label{app:linguistic}

This section analyzes the lexical and syntactic properties of synthetic data in detail to understand how linguistic variation contributes or limits model performance. 
While interpretability-guided generation enhances alignment with emotionally salient features, it also imposes constraints on linguistic breadth and expressive authenticity.

\subsection{Type-Token Ratio}

As shown in Table \ref{tab:ttr}, both SHAP-guided and naïve synthetic data exhibit lower lexical diversity than real data.
This reduced type-token ratio stems from the repetitive nature of LLM-generated content.
In SHAP-guided generation, such repetition appears strategic: it amplifies emotion-relevant signals emphasized by the model, which is particularly beneficial for minority classes like \textit{optimism}, where focused exposure to consistent, class-distinctive patterns is critical for learning.

\begin{table}[h]
\centering
\caption{SHAP-guided augmentation produces less lexically diverse text due to strong emphasis on emotionally discriminative patterns over vocabulary breadth.}
\label{tab:ttr}
\begin{tabular}{@{}l@{\hspace{3em}}c@{}}
\toprule
\textbf{Augmentation Strategy} & \textbf{Type-Token Ratio} \\
\midrule
Real Data Expansion           & 0.2405 \\
Naïve Generation              & 0.1428 \\
SHAP-Guided Generation        & 0.1332 \\
\bottomrule
\end{tabular}
\end{table}

However, this focused repetition also introduces a cost. 
Repetition may limit generalization that causes models to overfit to specific lexical constructions.
This risk is particularly pronounced in social media, where evolving slang and idiomatic variation are central to authentic emotional expression.
Notably, while naïve generation yields slightly higher lexical diversity than SHAP-guided text, this variety is unfocused, and its lack of alignment with emotion-relevant features leads to inferior classification performance, thus implying that lexical diversity alone does not guarantee utility; relevant linguistic markers remain paramount.

\subsection{Lexical Overlap}

To further assess how well synthetic text aligns with real text, we compute Jaccard similarity scores between real and generated datasets (Table \ref{tab:jaccard}).
SHAP-guided augmentation consistently achieves higher overlap across all emotions, suggesting that intepretability-informed prompts help maintain lexical fidelity.
Despite being the least frequent, the highest alignment appears in optimism, indicating that SHAP helps focus LLM output on meaningful emotional cues even under data scarcity.

\begin{table}[h]
\centering
\caption{Jaccard scores show that SHAP-guided augmentation better preserves emotion-specific vocabulary than naïve generation. The highest overlap in optimism suggests SHAP's strength under data scarcity, while lower alignment in joy reflects the challenge of modeling its stylistic variability with keywords and exemplars alone.}
\label{tab:jaccard}
\begin{tabular}{@{}l@{\hspace{3em}}c@{\hspace{1em}}c@{}}
\toprule
\textbf{Emotion} & \textbf{Real vs. SHAP-Guided} & \textbf{Real vs. Naïve} \\
\midrule
Anger    & 0.2090 & 0.1837 \\
Joy      & 0.1876 & 0.1705 \\
Optimism & 0.2470 & 0.1831 \\
Sadness  & 0.2316 & 0.2098 \\
\bottomrule
\end{tabular}
\end{table}

Conversely, the joy class shows the lowest overlap with SHAP-guided outputs, likely due to its lexical expression being more stylistically varied and harder to simulate through exemplars and keywords alone.
Naïve generation shows lower and more uniform overlap across emotions, reinforcing the finding that without interpretability guidance, lexical alignment to real-world emotional text is less consistent.

\subsection{Part-of-Speech Analysis}

\begin{figure*}[h]
% \centering
\includegraphics[width=\textwidth]{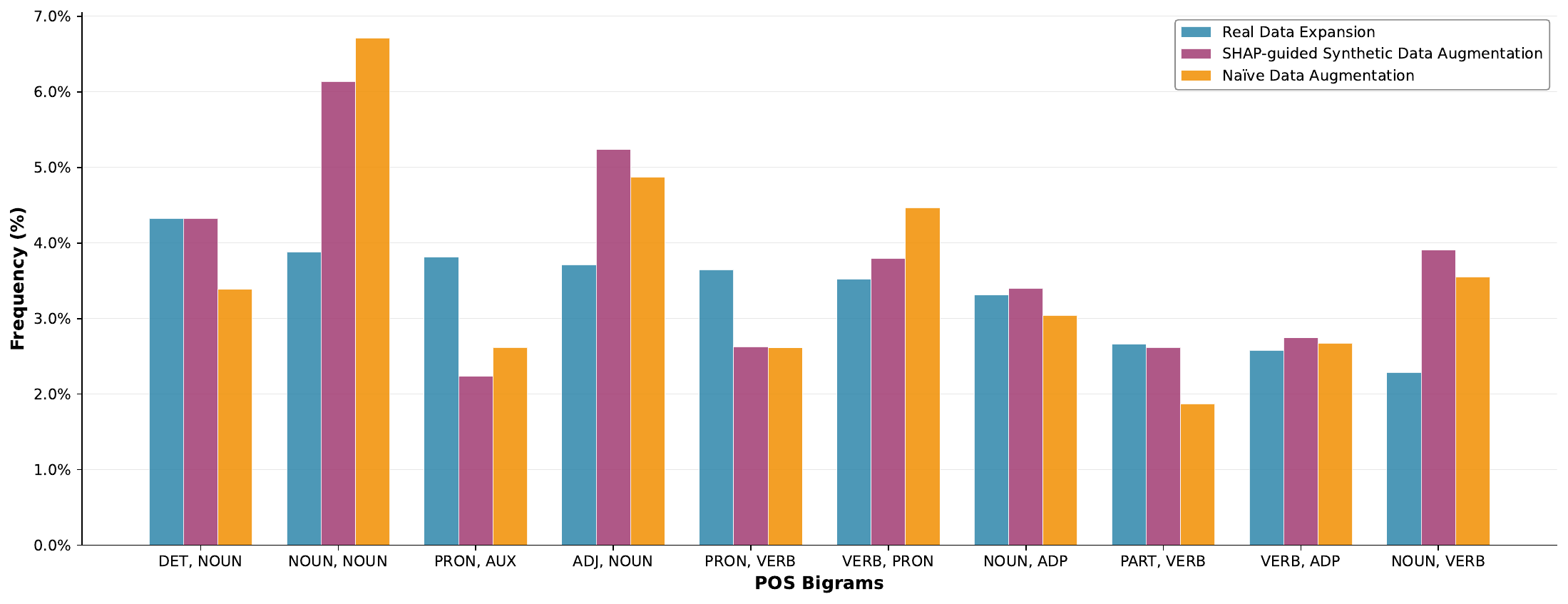}
\caption{POS bigram analysis shows SHAP-guided data mimics real syntactic patterns more closely than naïve data, especially in descriptive constructions, though both synthetic types reduce personal framing sequences.}
\label{fig:bigram_full}
\end{figure*}

Syntactic patterns further reveal differences between the SHAP-based and naïve strategies. 
As Figure~\ref{fig:bigram_full} shows, interpretability-guided augmentation more closely emulates real emotional language by emphasizing content-rich bigrams. 
It over-produces descriptive pairs like (ADJ, NOUN) and (NOUN, VERB), and even exceeds their frequency in real data expansion. 
These combinations convey clearer emotional cues, which benefit the emotion recognition model.
Qualitative examination reveals that SHAP-guided generation demonstrates higher usage of explicit emotional descriptors over naïve. Representative examples include:

\begin{itemize}
\item \textit{"This customer service is absolutely fucking horrible. Been on hold for 45 minutes and they just hung up on me \#rage \#customerservicefail"}
\item \textit{"The way they handled that situation was absolutely horrible. Someone needs to be held accountable \#grim \#offense"}
\item \textit{"I'm so done with these bullies thinking they can just walk all over everyone. Time for some revenge \#fuming \#angry"}
\end{itemize}

These constructions provide unambiguous classification signals through explicit descriptors and emotion-bearing hashtags.
By contrast, naïve generation relies more on situational description with implicit emotional expression:

\begin{itemize}
\item \textit{"Just spent 3 hours on hold with customer service only to be told they can't help me. ARE YOU KIDDING ME?? \#CustomerServiceFail \#Furious"}
\item \textit{"Traffic has been at a standstill for 45 minutes because of construction that isn't even happening right now (emoji) \#TrafficRage"}
\item \textit{"Just found out my 'best friend' has been talking trash about me behind my back for months. Fake people everywhere (emoji) \#TwoFaced"}
\end{itemize}

Here, anger is conveyed through capitalization, emoji, and contextual framing rather than explicit descriptors, reflecting a less targeted replication of SHAP-identified discriminative features.

However, both synthetic approaches underrepresent certain personal voice patterns that are prevalent in user-generated posts. 
Notably, they produce fewer bigrams involving pronouns (e.g. PRON-AUX as in "I am", or PRON-VERB as in "I feel"), indicating a reduction in first-person perspective and temporal framing. 
When personal framing does appear, it typically manifests as simple present-state declarations:

\textbf{SHAP-guided:}
\begin{itemize}
\item \textit{"I'm fucking livid right now"}
\item \textit{"I've been trying to reach you for weeks about this billing error"} (rare temporal complexity)
\end{itemize}

\textbf{Naïve:}
\begin{itemize}
\item \textit{"I'm about to lose my mind over here"}
\item \textit{"I've been on hold for 2 HOURS and you just hung up on me"} (rare temporal complexity)
\end{itemize}

Both strategies produce minimal temporal complexity (e.g., \textit{"I have been dealing with this for months"} or \textit{"I should have seen this coming"}), lacking the narrative depth characteristic of authentic emotional posts.
As a result, synthetic outputs from either method sound less personal or narrative-driven than real social media text, favoring impersonal constructions over experiential first-person narratives.

Additionally, the naïve method shows a distinct tendency to overuse successive nouns (NOUN-NOUN chains) far more than either real text or the SHAP-guided data.
This heavy noun chaining suggests a more uncontrolled, exploratory style of generation.
Indeed, as noted in our earlier lexical analysis, an LLM without interpretability guidance tends to produce diverse but less semantically grounded content. 

\begin{figure*}[h]
\centering
\includegraphics[width=\textwidth]{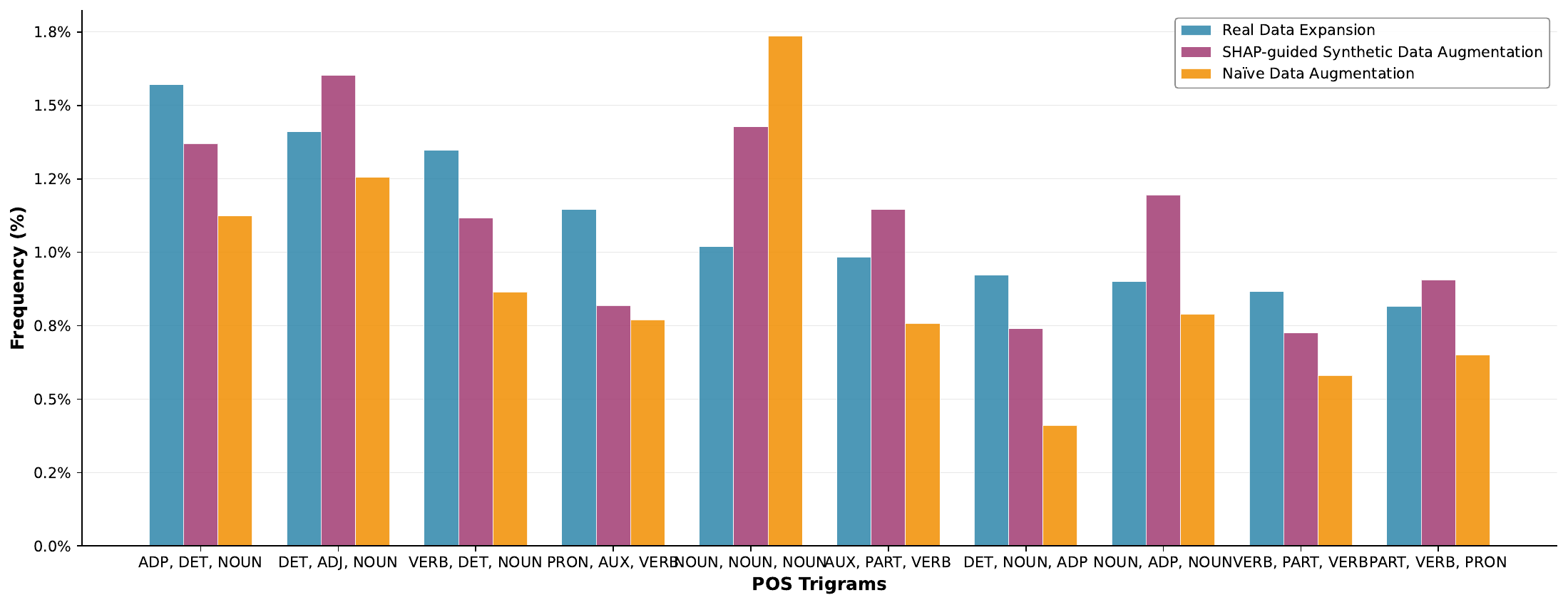}
\caption{Part-of-speech trigram analysis showing SHAP-guided data overproduces "DET, ADJ, NOUN" sequences, creating elaborate descriptive constructions that enhance classification but sacrifice authentic social media brevity.}
\label{fig:trigram}
\end{figure*}

The trigram analysis (Figure~\ref{fig:trigram}) further reveals that SHAP-guided data significantly overproduces "DET, ADJ, NOUN" sequences, creating more elaborate descriptive constructions that enhance classification signals but sacrifice the brevity characteristic of authentic social media expression. 
Examples of this pattern include:

\begin{itemize}
\item \textit{"The way some guys talk to women online is absolutely disgusting. Show some fucking respect \#harassment"}
\item \textit{"Can't believe my flight got delayed AGAIN for 3 hours with no explanation. Customer service is absolutely horrible \#fuming \#airlineproblems"}
\end{itemize}

These multi-layered descriptive constructions provide clear emotional classification cues but sound syntactically denser than authentic venting, which might simply state \textit{"Customer service sucks"} or \textit{"They hung up on me again"}.

Both synthetic approaches underproduce "PRON, AUX, VERB" patterns and "DET, NOUN, ADP" sequences, indicating reduced personal framing and prepositional complexity. 
Moreover, both strategies severely underutilize particle verbs crucial for expressing dynamic emotional states. 
The few instances found include:

\begin{itemize}
\item SHAP: \textit{"dont give up now, ur closer than u think! \#persistence"}
\item Naïve: \textit{"why do i always mess up the good things in my life?? hate myself rn"}
\end{itemize}

Meanwhile, constructions common in authentic posts—\textit{"break down"}, \textit{"freak out"}, \textit{"blow up"}, \textit{"piss off"}—were virtually absent.
This results in synthetic data that favors present-state declarations over the temporally complex constructions and action-oriented language typical of genuine emotional posts.

Together, these observations illustrate a fundamental trade-off in our augmentation strategy. 
On one hand, the SHAP-guided approach aligns synthetic data more closely with real emotional text by reinforcing the most discriminative lexical and syntactic features—as evidenced by examples like \textit{"absolutely fucking horrible"} and explicit emotion hashtags (\#fuming, \#outrage).
On the other hand, it achieves this alignment at the cost of some linguistic diversity and natural expressiveness, producing syntactically dense constructions that lack the variability, first-person narrative depth, and temporal complexity characteristic of authentic social media text.
Crucially, this targeted focus on salient patterns boosts emotion classification performance, proving particularly beneficial for minority classes where concentrated emotion-bearing terms compensate for limited training examples.
In short, SHAP-guided generation succeeds not by broadly mimicking all facets of natural language, but by amplifying exactly the features that matter most for emotion recognition—a strategy that improves model performance while revealing the linguistic authenticity constraints of current LLM-based synthetic data generation.

%% ====== APPENDIX: ABLATION STUDY ======
\section{Ablation Study: The Effect of Exemplars}
\setcounter{figure}{0}
\setcounter{table}{0}
\renewcommand{\thefigure}{H\arabic{figure}}
\renewcommand{\thetable}{H\arabic{table}}
\label{app:ablation}

\begin{figure*}[t]
\includegraphics[width=\textwidth]{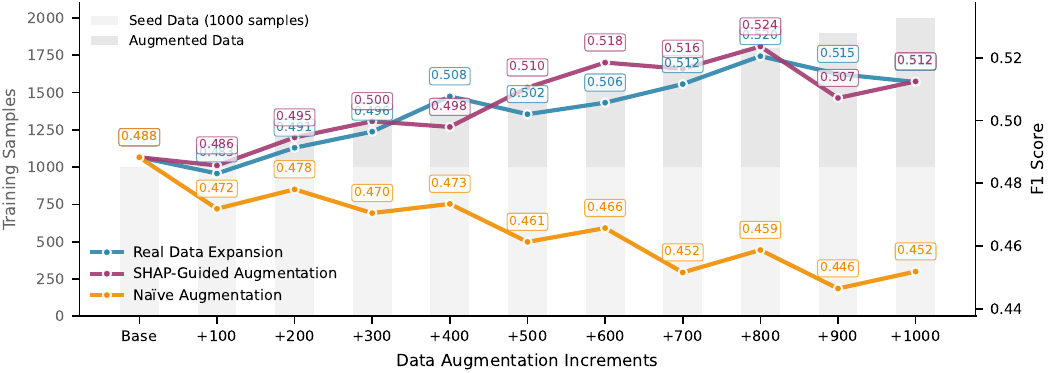}
\caption{Ablation study comparing SHAP-guided generation with and without exemplars. 
The exemplar-free approach loses its advantage over real data expansion, demonstrating that SHAP-derived keywords alone are insufficient without contextual and stylistic guidance from exemplars.}
\label{fig:ablation}
\end{figure*}

To isolate the contribution of exemplars to synthetic data quality, we compare SHAP-guided generation with and without exemplar guidance. Figure~\ref{fig:ablation} demonstrates that removing exemplars reduces performance from 0.538 to 0.526,  eliminating the characteristic advantage over real data expansion that  exemplar-informed generation maintains across increments.

These findings demonstrate that SHAP-derived keywords alone are insufficient for high-quality generation. Exemplars provide crucial contextual and stylistic information enabling LLMs to generate emotionally authentic text that effectively reflects SHAP-identified features. 
Notably, the exemplar-free SHAP strategy still maintains substantial advantage over naïve generation, indicating that
interpretability guidance provides value even without exemplars, but optimal performance requires both components.

\end{document}